\documentclass[letterpaper, 10 pt, conference]{ieeeconf}
\IEEEoverridecommandlockouts
\usepackage{amsmath,amssymb,amsfonts}
\usepackage{algorithmic}
\usepackage{graphicx}
\graphicspath{{figures/}}
\usepackage{textcomp}
\usepackage{xcolor}
\usepackage{svg}
\usepackage{subcaption}
\usepackage{booktabs}
\usepackage{url}
\usepackage{siunitx}
\usepackage{tikz}
\usepackage{soul}
\usepackage{pifont}
\usepackage{makecell}
\let\labelindent\relax
\usepackage{enumitem}
\usepackage{stfloats}
\usepackage[%
doi=false, mincitenames=1, maxcitenames=2, maxbibnames=99,isbn=false,url=false,eprint=true]{biblatex}
\usepackage[hidelinks]{hyperref}
\newcolumntype{?}{!{\vrule width 0.8pt}}
\AtNextBibliography{\small}

\def\BibTeX{{\rm B\kern-.05em{\sc i\kern-.025em b}\kern-.08em
    T\kern-.1667em\lower.7ex\hbox{E}\kern-.125emX}}
\begin{document}

\title{
4D Radar Semantic Segmentation of People in Field Conditions Using Temporal Multi-View Networks 

\thanks{The authors are with the Center for Advanced Autonomous Sensor Systems (AASS), Örebro University, Sweden. $^{*}$Corresponding author oleksandr.kotlyar@oru.se. 
This work has received funding from Sweden's Innovation Agency under grant agreement 2021-04714 (Radarize), 2025-01043 (CLEARPATH) and from the European Union's Horizon Europe Framework Programme under the RaCOON project (ID: 101106906). The computations and data handling were enabled by resources provided by the National Academic Infrastructure for Supercomputing in Sweden (NAISS), partially funded by the Swedish Research Council through grant agreement no. 2022-06725, projects NAISS 2025/5-748 and NAISS 2025/6-478.}
}

\author{Mikael Skog, Oleksandr Kotlyar$^{*}$, Vladim\'ir Kubelka and Martin Magnusson}

\maketitle

\begin{abstract}
    Reliable people detection is crucial for the safe autonomy of mobile robots and heavy vehicles, both on roads and in industrial settings like mining and construction.
    However, common sensors like cameras or lidars are prone to failure in adverse conditions such as dust, fog, or smoke, which limits their use in real-world robotic systems.
    Radar, on the other hand, delivers robust measurements in a wide range of environmental conditions.
    In particular, modern high-resolution 4D imaging radars provide 4D point clouds across range, azimuth, and elevation, as well as per-point Doppler velocity data, well suited for robot perception.  
    We propose TMVA4D, a family of artificial neural network architectures based on CNN and ConvLSTM encoders that leverage the 4D radar modality for semantic segmentation.
    The architectures are trained to distinguish between background and person classes using a series of 2D projections of the 4D radar data, encompassing elevation, azimuth, range, and Doppler velocity dimensions.
    Evaluated across several operational sites, our models achieve promising performance (Dice 75.9\%, IoU 61.2\% for class person) even in low-visibility conditions.
    The data and code will be made publicly available upon publication.
    
\end{abstract}

\section{Introduction}
\begin{figure}[t]
    \centering
    \begin{subfigure}[t]{0.235\textwidth}
        \centering
        \includegraphics[width=\textwidth, height=\textwidth]{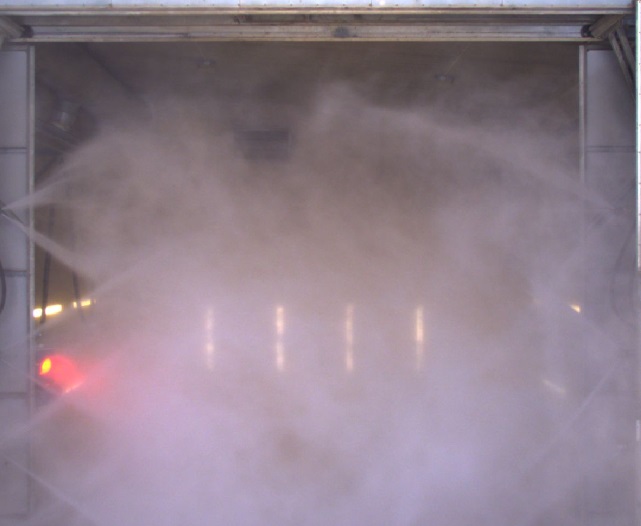}
        \vspace{-15pt}
        \caption{Color camera image}
        \label{fig:front-color}
        \vspace{8pt}
    \end{subfigure}
    \begin{subfigure}[t]{0.235\textwidth}
        \centering
        \includegraphics[width=\textwidth]{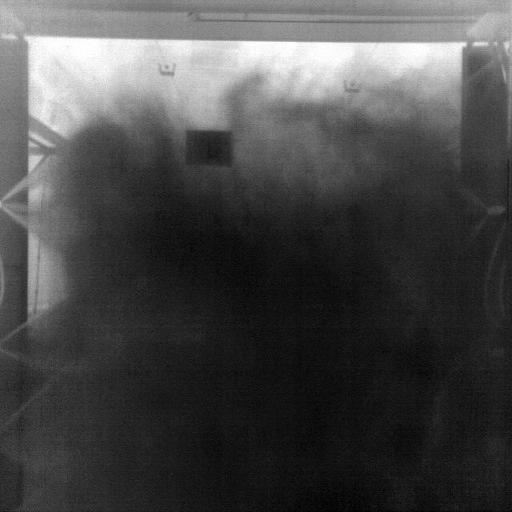}
        \vspace{-15pt}
        \caption{Thermal camera image}
        \label{fig:front-thermal}
    \end{subfigure}
    \begin{subfigure}[t]{0.235\textwidth}
        \centering
        \includegraphics[width=\textwidth]{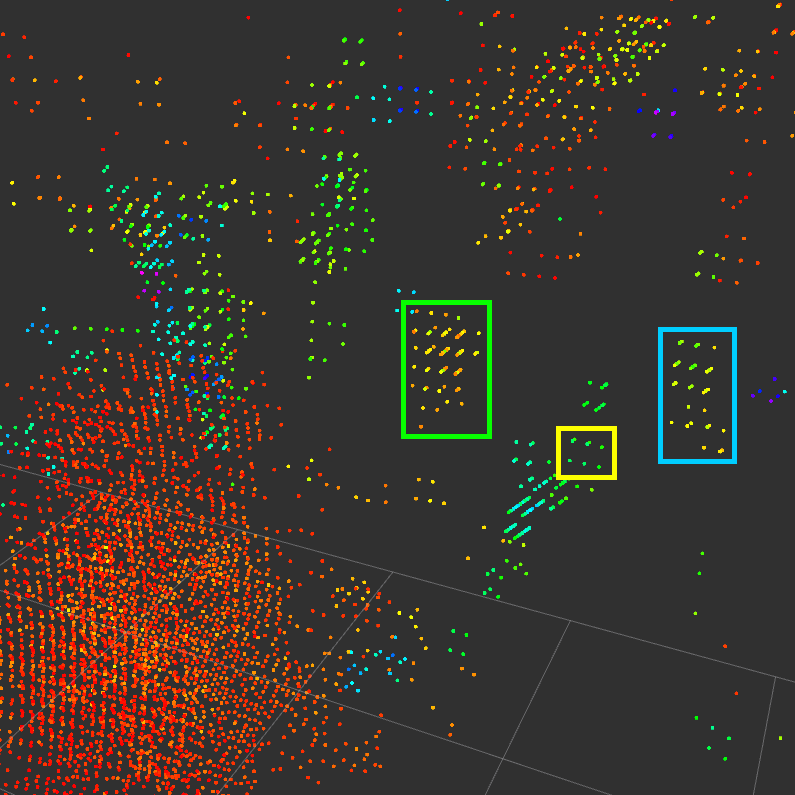}
        \vspace{-15pt}
        \caption{Radar point cloud}
        \label{fig:front-pcd}
    \end{subfigure}
    \begin{subfigure}[t]{0.235\textwidth}
        \centering
        \includegraphics[width=\textwidth]{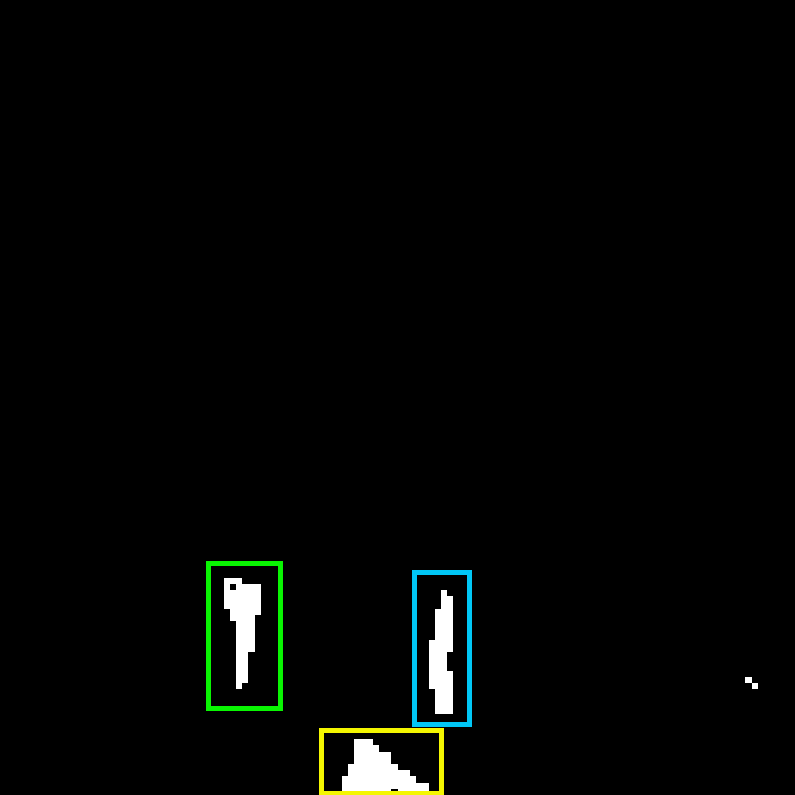}
        \vspace{-15pt}
        \caption{Prediction in camera view}
        \label{fig:front-pred}
    \end{subfigure}
    \caption{Indicative results from our semantic segmentation pipeline for human detection, showing people in an industrial car wash.
    A ``wall'' of sprayed water obscures the view for both the color camera (\subref{fig:front-color}), thermal camera (\subref{fig:front-thermal}), and lidar (not shown), rendering the people undetectable.
    A 4D radar still records data from behind the water (\subref{fig:front-pcd}). 
    The output of our proposed TMVA4D architectures is a predicted mask in the camera (elevation-azimuth) view (\subref{fig:front-pred}). The mask depicts the \texttt{background} class in black and \texttt{person} class in white. Colored bounding boxes (for visualization only) indicate the same person in the raw radar point cloud and prediction output.}
    \label{fig:front}
    \vspace{-0.4cm}
\end{figure}

For an autonomous vehicle to safely navigate complex environments, accurate perception and internal representation of its surroundings is critical. In particular, navigating spaces shared with people requires reliable human detection. To this end, vision-based systems and lidar sensors can provide the vehicle with low-noise, high-resolution data under ideal visibility conditions. However, they both fall short in adverse conditions such as rain, fog, dust, or smoke~\cite{lidar-bad-dawson, lidar-bad-kim}. Unlike cameras and lidars, modern radar sensors enable robotic perception of objects otherwise obscured by precipitation and airborne particles (see Fig.\ref{fig:front}), albeit with the disadvantages of lower angular resolution and more noise.
This makes radar a compelling option for detecting people under low-visibility conditions.
This is particularly important in field robotics applications such as mining, construction, and forestry, where adverse visibility is common and operational safety is imperative.

Advancements in deep learning have established it as the dominant approach for object detection and segmentation across radar and other sensing modalities. Previous research \cite{ramp-cnn,
mvrss,rodnet,RadarVelocityTransformer} on segmentation of two-dimensional radar data has focused on radars providing range-azimuth (RA) or range-azimuth-Doppler (RAD) information.
While such data can be obtained with relatively inexpensive radar technology, these sensors lack an elevation dimension.
Conversely, radars providing power measurements in three spatial dimensions (range, azimuth, elevation) plus Doppler are also available and are often denoted as 4D, or 3+1D, radars. The elevation data may prove critical in cases where people %
have collapsed due to loss of consciousness, since motionless targets are indistinguishable through Doppler measurements.

We present TMVA4D %
-- a family of temporal multi-view neural networks with atrous spatial pyramid pooling (ASPP) modules~\cite{aspp} designed for processing 4D radar data -- that leverage 2D and 3D convolutional (CNN) as well as convolutional long short-term memory (ConvLSTM)~\cite{shi2015} neural networks for human detection. %
TMVA4D is a deep neural network (DNN) tailored for semantic segmentation of people using sets of 2D heatmaps projected from 4D radar point clouds. For TMVA4D, we evaluate the performance of several encoders composed of different combinations of CNN and ConvLSTM layers.

To train and evaluate TMVA4D models, we introduce a novel dataset. The annotation and formatting process is presented in the paper. 
By evaluating the performance of TMVA4D models on the dataset, we show that our approach to representing and segmenting people in radar data is effective, even where other sensor modalities fail due to low-visibility conditions (as shown in Fig.~\ref{fig:front}). %
Beyond training and evaluation, we validate the method through deployment in a live system, highlighting its applicability for real-time perception and safe autonomy in field robotics environments.

In summary, our contributions are as follows:
\begin{enumerate}
\item a family of architectures for semantic segmentation from heatmaps projected from 4D radar point clouds, tailored for robotic perception in adverse conditions;
\item a multimodal dataset (4D radar and thermal camera) for training and evaluating people detection in 4D radar heatmaps including challenging conditions from mining and construction use cases;
\item demonstrating state-of-the-art performance in semantic segmentation of people, based on a thorough evaluation, with live validation highlighting its applicability for real-time robotic systems.
    
\end{enumerate}

\section{Related Work}
\label{sec:related}
\subsection{2D datasets and methods}
As 2D radar is a well-established modality, there are several datasets available for people detection.
The CARRADA dataset~\cite{carrada} comprises data from a stationary automotive radar in a flat and open environment. It represents the radar data as heatmaps in polar coordinates in three views: range-azimuth (RA), range-Doppler (RD), and azimuth-Doppler (AD). Annotations for multiple types of moving objects are provided in different forms, including semantic segmentation masks.
The CRUW dataset~\cite{cruw} contains automotive radar data from on-road driving  with varying levels of illumination. The radar data is represented as RA heatmaps in a Cartesian BEV, with point-wise annotations for object instances.
The RADDet dataset~\cite{raddet} includes urban automotive radar data represented as RA and RD heatmaps with annotations in the form of 2D bounding boxes.
RadarScenes~\cite{radarscenes} delivers 100 km of automotive radar data collected in various urban and rural scenarios. The data is represented as radar point clouds, with point-wise semantic and single object annotations.

There are also several deep learning-based architectures for detection and segmentation in 2D radar data.
Gao et al.~\cite{ramp-cnn} propose RAMP-CNN, a CNN architecture for object detection in the RA view. From RAD tensors, it extracts sequences of heatmaps in the RA, RD, and AD views which are processed by convolutional autoencoders in each view, followed by an inception module~\cite{googlenet}.
Wang et al.~\cite{rodnet} propose RODNet-HGwI, a 3D convolutional autoencoder architecture with inception modules in the decoder. It performs RA view object detection taking RA maps as input.
Ouaknine et al.~\cite{mvrss} and Decourt et al.~\cite{RECORD} propose the encoder-decoder architectures TMVA-Net and RECORD for semantic segmentation in the RA and RD views. Both TMVA-Net~\cite{mvrss} and RECORD~\cite{RECORD} take multiple frames of input in the RA, RD, and AD views, each view with its own encoder. TMVA-Net is a convolutional architecture with ASPP modules, while RECORD uses both CNN and ConvLSTM layers.
Zeller et al. present Radar Velocity Transformer~\cite{RadarVelocityTransformer} and Gaussian Radar Transformer~\cite{zeller-2023-transformer}, transformer-based architectures for moving object segmentation on noisy radar point clouds; using a U-Net structure, with transformer layers for feature extraction and transformer-based upsampling layers. 

\subsection{3D datasets and methods}

As we aim to detect people in various poses, including on the ground, 3+1D radar data is more suitable than 2D (or 2+1D).
Since 4D radar is a more recent modality, fewer human detection methods and datasets exist in the literature.

\textcite{k-radar} present the on-road K-Radar dataset with a network for object and human detection directly on 4D radar tensors. Not all 4D radars provide the full tensor as output, which is why we chose to use heatmaps computed from a 4D point cloud instead.
The View-of-Delft (VoD) dataset \cite{vod} provides 4D radar data collected in urban traffic. Annotations are provided as 3D bounding box of both moving and static objects, including pedestrian, cyclist, and car objects.
\textcite{tj4dradset} present TJ4DRadSet (TJ4D), an automotive dataset with 4D radar data. They include human detection results using an adaptation of PointPillars~\cite{pointpillars} on the radar point clouds. %
RadarPillars~\cite{radarpillars} presents further adaptation of PointPillars made to handle the sparsity of radar point clouds, demonstrating more accurate bounding box detections in the VoD dataset.
RadarNeXt~\cite{radarnext} improves on the pillarization approach for radar point clouds, using Reparameterizable Depthwise Convolutions to significantly reduce the number of network parameters. However, the methods proposed in \cite{tj4dradset}, \cite{radarpillars}, and \cite{radarnext} output bounding boxes, while our method performs semantic segmentation. 
ImmFusion~\cite{immfusion} recovers full body poses from 4D radar data fused with camera images.
RAPTR~\cite{raptr} estimates 3D poses from ER and RA projections of 4D radar data leveraging transformer decoders. 
In constrast to \cite{immfusion} and \cite{raptr} which both focus on pose recovery of nearby humans, our work focuses on people detection in difficult field conditions including ones at a distance.
Closer to the use case targeted in this paper, \textcite{wang-2024-fusion} propose a graph neural network-based model for object detection and semantic segmentation of 4D radar point clouds.

To the best of our knowledge, this paper presents the first study of 2D heatmap-based semantic segmentation of 4D radar data collected in adverse conditions.
Unlike all aforementioned datasets, our dataset includes challenging field conditions relevant to mining, construction, and forestry, with visibility reduced by induced dust, water mist/spray, and smoke.
In contrast to prior human detection methods, TMVA4D has been trained on people engaged in various activities, such as crouching, running, and climbing.

\section{Dataset}
\subsection{Data Collection}\label{AA}
\begin{figure}
    \centering
    \begin{subfigure}[t]{0.235\textwidth}
        \centering
        \includegraphics[width=\textwidth]{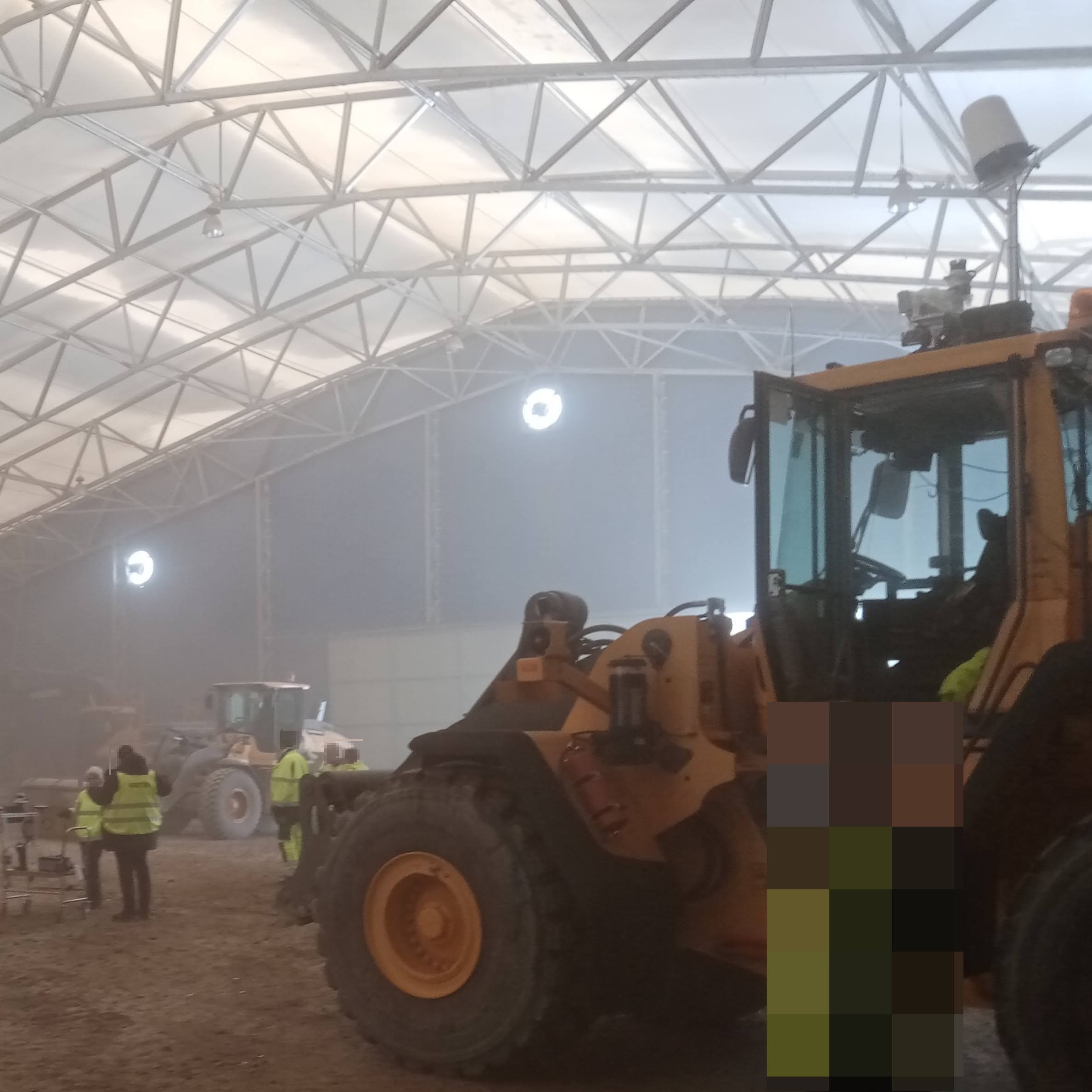}
    \end{subfigure}
    \begin{subfigure}[t]{0.235\textwidth}
        \centering
        \includegraphics[width=\textwidth]{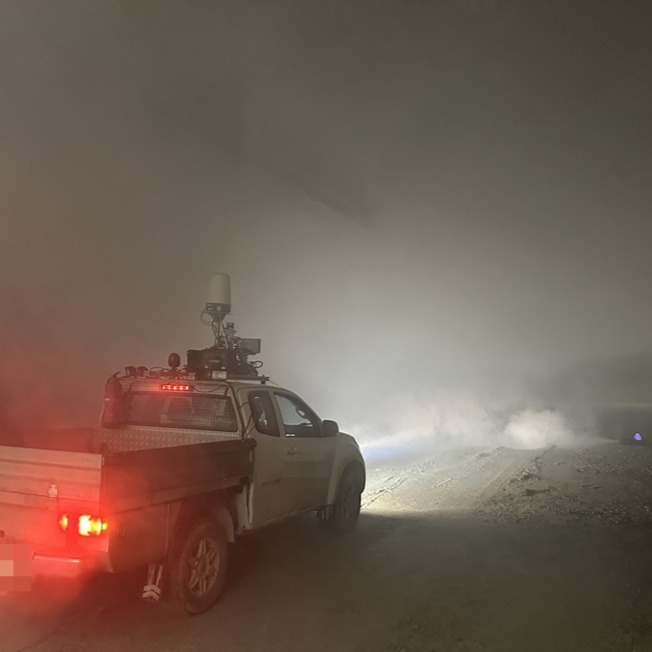}
    \end{subfigure}
    \caption{The platforms used for recording the data set used for real-world evaluation. \textit{Left:} wheel loader in a dusty environment, as the dust settles. \textit{Right:} mine car in smoke-filled tunnel.}
    \label{fig:platforms}
    \vspace{-0.4cm}
\end{figure}

Our dataset was collected %
using a vehicle-mounted sensor system in four environments: an underground mine, a large car wash, %
an industrial tent used in testing of construction vehicles, and an outdoor wooded area. 
A large portion of the data features people in different positions and performing various actions, including lying down, standing still, walking, running, and climbing. Certain sequences feature a dummy in place of an actual person. In several sessions, visibility was reduced with airborne particles induced to replicate realistic and challenging scenarios: airborne dust, water spray, and smoke. Part of the data was collected during ego-motion.
Fig.~\ref{fig:platforms} shows the two platforms used, in two of the scenarios.

Data collection was carried out using a thermal imaging camera (FLIR AX5) and a 4D solid-state radar (Sensrad Hugin A3-Sample). The camera captured images used to annotate the radar data for training and testing of the TMVA4D models. The data collected by the radar are represented as point clouds and we chose to trade sensor range (\SI{42}{\meter}) for depth resolution (\SI{0.1}{\meter}). The data was recorded using ROS~1 as time\-stamped images and Point Cloud Data (PCD) files.

\subsection{Annotation Process}
We provide ground-truth segmentation masks in the 2D azimuth-elevation (EA) projection of the radar data, generated using the thermal camera data. %
To produce these labels, the thermal images were first scaled from a resolution of $640\times512$ to $512\times512$.
4,900 (4.8\%) of the 102,966 recorded images were manually annotated in CVAT\footnote{\url{https://www.cvat.ai/}} using polygon annotation to separate people from the background.

The manually annotated thermal images were used to train a custom YOLOv8\footnote{\url{https://yolov8.com/}} model for automatic annotation.
The annotated images were randomly split into training and validation sets with a 90:10 split ratio. To decrease the training time and improve generalization, training was performed on the pre-trained segmentation model YOLOv8x-seg.
Our trained YOLOv8 model achieves a segmentation performance of 97.84\% mAP50 and 80.19\% mAP50-95, with 96.95\% precision and 96.64\% recall on the validation set. The validation set was only used to determine the number of epochs required for convergence, it was not used to tune any other hyperparameters. These results indicate that our automatically generated annotations are highly accurate, which was further supported by qualitative inspection. %

We use this model to semantically segment all resized images for class \texttt{person}.
These segmentation masks constitute the training labels for our TMVA4D models, since the radar heatmaps used by TMVA4D result from point cloud projections to the coordinate frame of the thermal camera.
To reduce the TMVA4D architecture sizes and thereby lower inference times, masks are resized from $512 \times 512$ to $128\times128$, with regard to the density of the point clouds.
\subsection{Radar Data Representation}\label{sec:representation}
From each 4D radar point cloud, a heatmap is generated in each of five separate views with different dimensions:
\begin{itemize}
    \itemsep0em
    \item Elevation-azimuth \,(EA) -- $128\times128$.
    \item Elevation-range \ \ \ \,\hspace{-0.5pt}(ER\hspace{0.6pt})\hspace{-0.1pt} -- $128\times256$.
    \item Elevation-Doppler \,\hspace{-0.5pt}(ED)\hspace{-0.15pt} -- $128\times256$.
    \item Range-azimuth \ \ \ \ (RA\hspace{-0.8pt}) -- $256\times128$.
    \item Doppler-azimuth \ \ (DA\hspace{-0.9pt}) -- $256\times128$.

\end{itemize}
Each heatmap in the EA view can be conceptualized as a uniform grid spanning the camera frame, extending slightly beyond the image boundaries to align the grid with the distribution of the projected points. Considering the dimensions separately, the azimuth and elevation dimensions of this grid both span an interval of values, corresponding to positions on the camera frame along its width or height. We aim at partitioning each of these intervals into 128 bins.
However, due to the sparsity of the input points, the elevation and azimuth dimensions are first discretized into 28 and 44 bins, respectively.
After all cells are assigned values using the process described below, these dimensions are upscaled in all views to 128 bins using linear interpolation.
The range and Doppler dimensions are discretized into 256 bins each.

For a heatmap in a given view, consider a point that falls within the $i^\text{th}$ bin of the vertical dimension and within the $j^\text{th}$ bin of the horizontal dimension when projected to the view. This point will then belong to the cell at row $i$ and column $j$ of the grid representing this heatmap.
Each cell is assigned a value based on all the points it contains.
The Sensrad Hugin radar produces point clouds where each point bears a power value expressed in dB and a Radar Cross Section (RCS) value expressed in dBsm (a logarithm of m$^2$).
We have evaluated applying the \textit{sum}, \textit{mean}, or \textit{maximum} function to either the power or RCS values of the points belonging to each cell.
We have evaluated the difference between the power and RCS (see Section~\ref{sec:training_testing}, Table~\ref{table:TMVA4D_CNN_performance}) by training and testing the TMVA4D-CNN-5 (see Section~\ref{sec:TMVA4D_Architectures}) architecture.
Based on the results, we have decided to limit the evaluation of the rest of the architectures to power values only (\emph{Max power}, \emph{Mean power}, and \emph{Sum power}). Fig.~\ref{fig:proj-ea-example} shows points projected to the EA view for one thermal image, with corresponding EA heatmap (maximum power values used for each cell).

\subsection{Dataset Composition}
Longer sequences have been split into shorter segments, with dataset size reduced by excluding segments without people. %
A frame in a sequence refers to heatmaps in all views originating from a singular point cloud, with annotations generated from its temporally closest thermal image. In the segmentation masks, evaluated across all frames in the dataset, the percentage of mask pixels representing class \texttt{person} is 1.1\%; and 61\% of the segmentation masks contain at least one pixel representing the class. The dataset consists of 67,345 frames, which are split into 56.3\% training, 16.5\% validation, and 27.2\% test parts. The file structure of our dataset follows that of the CARRADA dataset~\cite{carrada}.

\section{TMVA4D Architectures}
\label{sec:TMVA4D_Architectures}
\begin{figure}
    \centering
    \includegraphics[width=\linewidth]{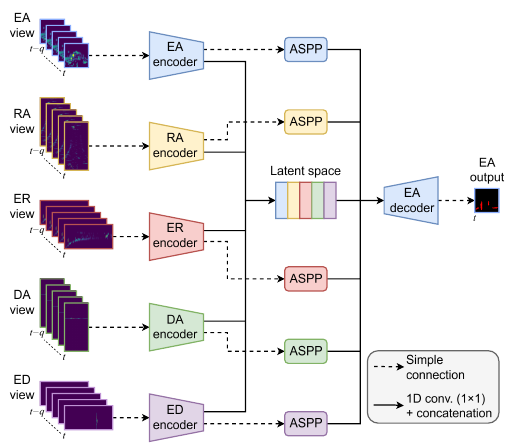}
    \vspace{-0.55cm}
    \caption{Overview of our proposed TMVA4D architectures. The input data is provided in the EA, ER, ED, RA, and DA views, the output predictions are in the %
    EA view. Heatmaps in the input views at time $t$ and of the $q$ previous frames are used to output predictions for the frame at time $t$. These predictions are segmentation masks of the output view, representing classes \texttt{background} and \texttt{person}.}
    \label{fig:tmva4d}
\end{figure}

\begin{figure}
  \centering
  \begin{tikzpicture}[]
    \node[anchor=south west,inner sep=0] (image) at (0,0)
    {\includegraphics[width=.85\linewidth]{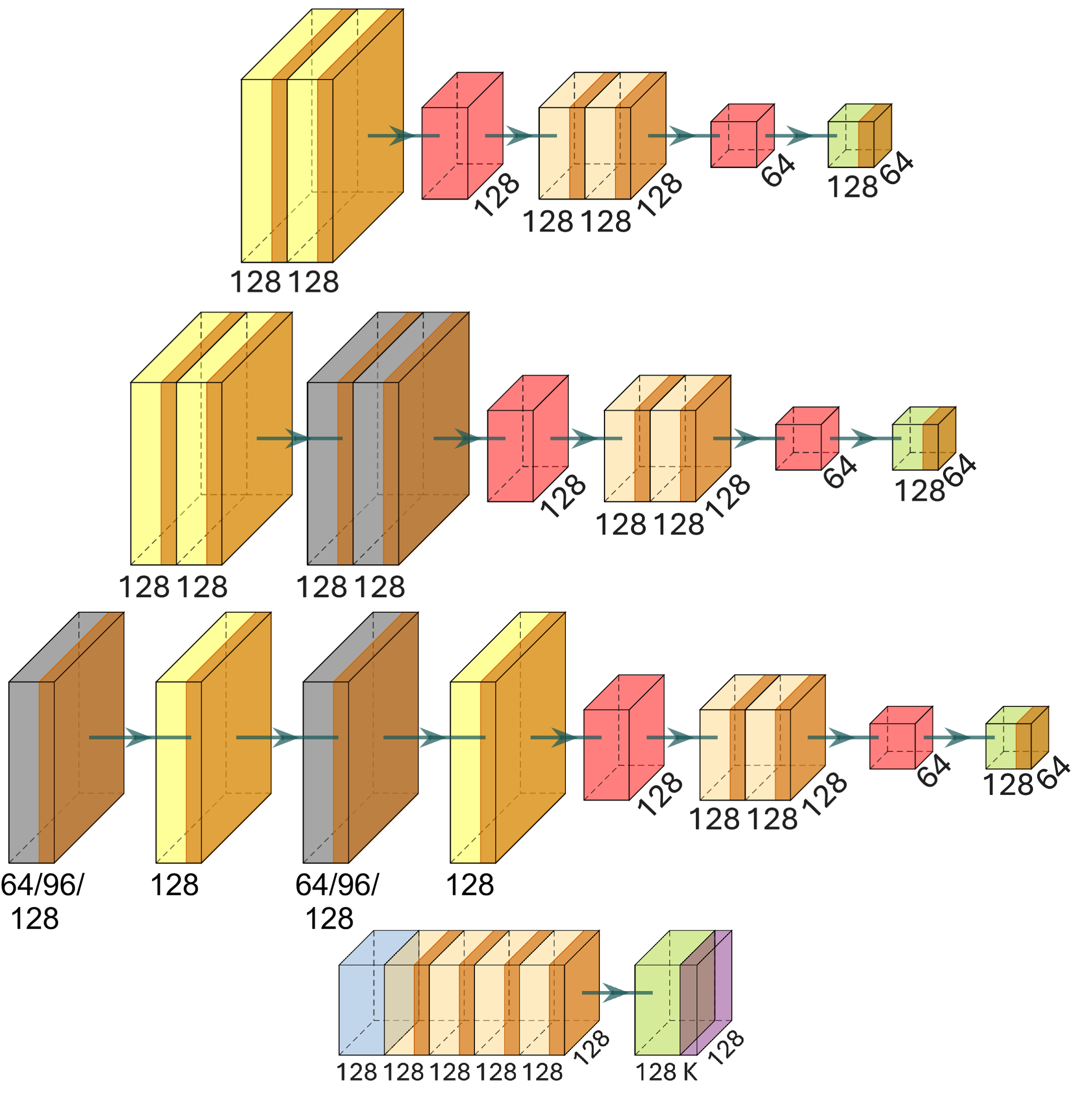}};
 
    \begin{scope}[x={(image.south east)},y={(image.north west)}]
        \draw (-0.05,.88) node[anchor=east] {(a)};
        \draw (-0.05,.6)  node[anchor=east] {(b)};
        \draw (-0.05,.34) node[anchor=east] {(c)};
        \draw (-0.05,.11) node[anchor=east] {(d)};  
    \end{scope}
  \end{tikzpicture}
    \begin{subfigure}[c]{0.43 \textwidth}
        \centering
        \includegraphics[width=\linewidth]{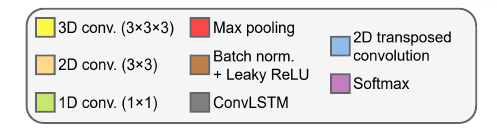}
        \phantomcaption
    \end{subfigure}
   \caption{TMVA4D encoder and decoder structures. (a) TMVA4D-CNN-$\{ i \}$ encoder: Input data is processed through a sequence of 3D and 2D convolutional layers for spatiotemporal feature extraction. The TMVA4D-CNN-5 encoder is shown in the picture. (b) TMVA4D–ConvLSTM encoder: Two ConvLSTM layers are added to the TMVA4D-CNN encoder, positioned between the stacks of 3D and 2D convolutional layers. (c) TMVA4D-IntConvLSTM encoder: Constructed with ConvLSTM layers interleaved through the 3D convolutional layers of the TMVA4D-CNN-5 encoder. (d) Decoder. Horizontal numbers indicate the number of filters used in each layer. Slanted numbers denote the width (and height) of the feature maps produced. \textit{K} represents the number of classes.
   }
   \vspace{-0.3cm}
   \label{fig:tmva4d-components}
\end{figure}

Our TMVA4D (temporal multi-view network with ASPP modules for 4D radar data) family of architectures are designed to take 4D radar data as input, formatted as multiple frames of heatmaps in the EA, ER, ED, RA, and DA views, as described in Sec.~\ref{sec:representation}.
The architectures predict semantic segmentation masks in the EA view for two classes: \texttt{background} and \texttt{person}. An overview of the general architecture is shown in Fig.~\ref{fig:tmva4d}, with its components and variants described in Fig.~\ref{fig:tmva4d-components}. 
TMVA4D architectures are inspired by TMVA-Net~\cite{mvrss}.
To leverage the performance of TMVA4D in capturing temporal dependencies of sequences of input data, we propose three variations of TMVA4D using different types of encoders, shown in Fig.~\ref{fig:tmva4d-components}a--c, namely: %
\begin{figure}[t]
    \centering
    \hspace*{-15pt}
    \begin{subfigure}[t]{0.24\textwidth}
        \centering
        \begin{tikzpicture}[white,font=\scriptsize]
            \draw[->](0,0) node[below]{0} -- (.85\textwidth,0) node[below]{128} --++ (0.4,0) node[below]{$j$};
            \draw[->](0,0)
            node[left=1mm,rotate=90,anchor=base]{0} -- (0,.85\textwidth) node[left=1mm,rotate=90,anchor=base]{128} --++ (0,0.4) node[left]{$i$};
        \includegraphics[width=.85\textwidth, height=.85\textwidth]{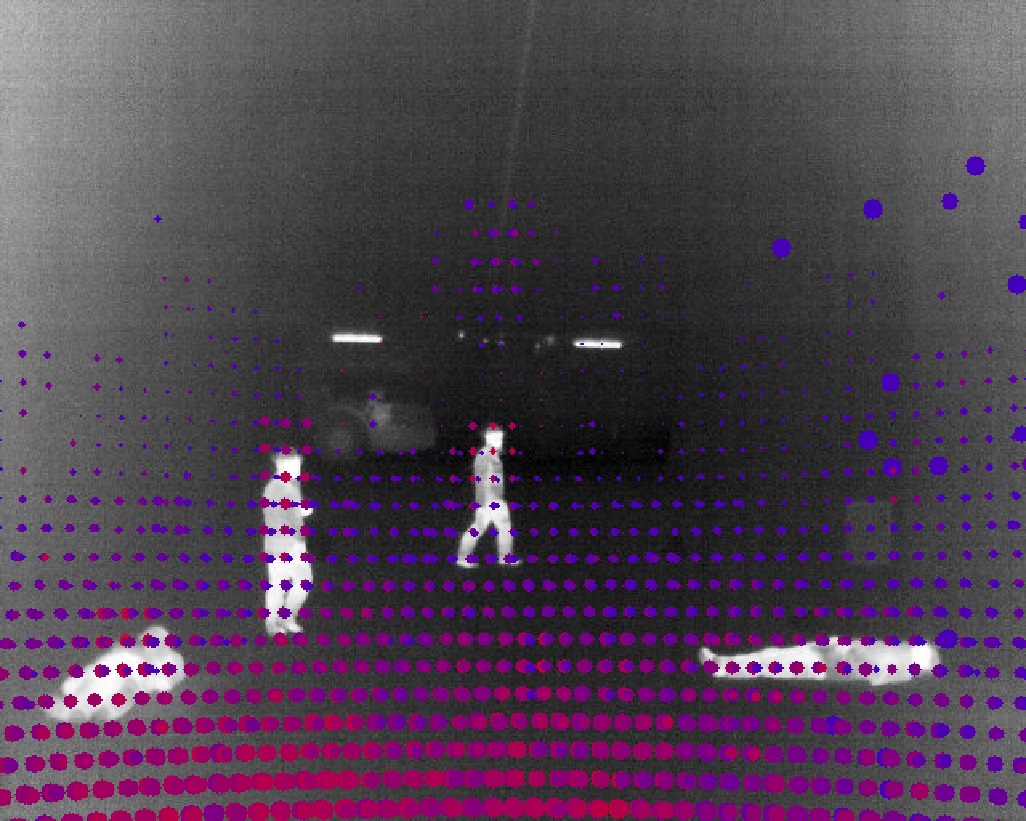}
        \end{tikzpicture}
        \vspace{-0.7cm}
        \subcaption{Points projected on image}
        \label{fig:proj-example}
    \end{subfigure}
    \begin{subfigure}[t]{0.24\textwidth}
        \centering
        \begin{tikzpicture}[font=\scriptsize]
            \draw[->](0,0) node[below]{0} -- (.85\textwidth,0) node[below]{128} --++ (0.4,0) node[below]{$j$};
            \draw[->](0,0)
            node[left=1mm,rotate=90,anchor=base]{0} -- (0,.85\textwidth) node[left=1mm,rotate=90,anchor=base]{128} --++ (0,0.4) node[left]{$i$};
            \includegraphics[width=.85\textwidth]{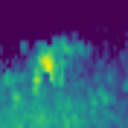}
        \end{tikzpicture}
        \vspace{-0.7cm}
        \subcaption{Elevation-azimuth heatmap}
        \label{fig:ea-proj-example}
    \end{subfigure}
    \hfill
  \caption[Point cloud projected to the EA view with corresponding EA heatmap]{Point cloud projected to the EA view with corresponding EA heatmap. (\subref{fig:proj-example}) Point cloud projected to the EA view, with brighter points indicating higher signal intensities, overlaid on top of the point cloud's temporally closest thermal image. (\subref{fig:ea-proj-example}) EA heatmap generated from point cloud using \textit{maximum} power values. Brighter colors represent higher values in the EA heatmap (viridis colormap).
}
  \label{fig:proj-ea-example}
  \vspace{-0.4cm}
\end{figure}
\vfill
\vfill

\begin{enumerate}[label=(\alph*)]
\item TMVA4D-CNN-$\{ i \}$, where $i \in \{1,3,5,7\}$ is the number of input frames (time steps). The input sequences of heatmaps are first processed by a set of 3D CNN layers capturing spatiotemporal features by performing convolutions across both spatial and temporal input dimensions. The number of 3D CNN layers is set to 1, 2, or 3 for models using 3, 5, or 7 input time frames, respectively. For TMVA4D-CNN-1, a single 2D CNN layer is used in place of 3D CNN layers.

\item TMVA4D–ConvLSTM. Data is processed by two 3D CNN layers followed by two ConvLSTM layers. The 3D CNN layers extract spatiotemporal features, 
capturing local patterns and motion dynamics in the input sequence. The ConvLSTM layers further model temporal dependencies, combining convolutions with the recurrent behavior of LSTMs to retain spatial information.

\item TMVA4D-IntConvLSTM. The encoder processes data by alternating ConvLSTM and 3D CNN layers. The first ConvLSTM layer captures the temporal dependencies from the raw input sequence while preserving spatial structure. The following 3D CNN layer extracts spatiotemporal features by convolving across spatial and temporal dimensions. The second Conv\-LSTM layer integrates temporal dynamics into the learned representation, while the final 3D CNN layer further refines the spatial and temporal feature representation. This design enables hierarchical spatiotemporal feature extraction with sequential temporal refinement.

\end{enumerate}

 At the end of all encoders, two 2D CNN layers are placed to capture finer spatial patterns, and finally a $1\times 1$ CNN block encodes temporal dynamics into a compact feature map to produce the final representation for further processing by TMVA4D. 
 Each encoder of TMVA4D takes a single channel input consisting of the heatmaps at time $t$ and of the previous $q$ frames, stacked depthwise. %
 Each 3D CNN layer reduces the depth of the feature maps by 2. 
 The encoders are connected to the decoder both directly and through multiple ASPP~\cite{aspp} modules, as shown in Fig.~\ref{fig:tmva4d}. Each ASPP module performs parallel convolutions at various dilation rates. We base our implementation on publicly available PyTorch code~\cite{mvrss}.%

\section{Model Training and Testing}
\label{sec:training_testing}
We trained the TMVA4D models on NVIDIA A100 (40~GB and 80~GB VRAM) and A40 (48~GB VRAM) GPUs.
All models were trained for 36 epochs,
allowing the models to converge on the validation set while preventing overfitting from additional training. The learning rate was initialized to $10^{-4}$ and multiplied by a decay factor of 0.9 after every two epochs.
The number of input frames (consecutive frames processed to generate predictions for the last) was set to one, three, five (as in~\cite{mvrss}), or seven for TMVA4D-CNN models, and to five for the networks with ConvLSTM encoder layers.

The number of filters in the 2D/3D CNN layers of the decoder and of all proposed encoders is set to 128. The TMVA4D–ConvLSTM encoders use the same number of filters.  %
For the TMVA4D-IntConvLSTM architecture, we tested three variants differing in the number of filters in the ConvLSTM layers. The ``Light" model utilizes 64 filters, while the ``Medium" and ``Heavy" models employ 96 and 128 filters in each ConvLSTM layer, respectively. Increasing the number of filters results in an increase of the number of trainable parameters from 65M through 97.7M to 132M, resulting in higher VRAM demands. To train the ``Heavy" model, we employed an NVIDIA A100 GPU with 80~GB  VRAM.  All hyperparameters were selected empirically. We set the batch size to 6 when training all models, except ``Heavy" TMVA4D-IntConvLSTM, for which we increased the batch size to 10 to achieve reasonable training times. %

For data augmentation, vertical and horizontal flipping is performed to mitigate overfitting. Data is normalized to the range [0, 1] using global dataset statistics, where minimum and maximum values across all heatmaps in each view are used to perform min-max normalization on the values of each view separately. The model parameters are updated using the ADAM optimizer~\cite{adam} for stochastic gradient descent.

To train the TMVA4D models, we use weighted cross-entropy loss $\mathcal{L}_\text{wCE}$ (\ref{eq:wCE_loss}), soft Dice loss $\mathcal{L}_\text{SDice}$ (\ref{eq:SDice_loss})~\cite{sdice}, or their combination for the EA view (\ref{eq:loss}). %
Consider the per-class binary segmentation masks for a given frame. In the EA view with dimensions $M \times N = 128 \times 128$, the annotated mask for class $c$ is a matrix $Y^{(c)} = (y_{ij}^{(c)})_{1 \leq i \leq M, 1 \leq j\leq N} $ of binary labels, with the corresponding prediction matrix $P^{(c)}= (p_{ij}^{(c)})_{1 \leq i \leq M, 1 \leq j\leq N} $ of probabilities $p^{(c)} \in [0, 1]$.
With $K$ classes $C = \{c_1, ..., c_K\}$, the weighted cross-entropy loss for this frame is given by 
\begin{equation}\label{eq:wCE_loss}
    \mathcal{L}_\text{wCE}=-\frac{1}{K}\sum_{c \in C} w_c \sum_{i=1}^M\sum_{j=1}^N y_{ij}^{(c)} \text{log}(p_{ij}^{(c)}) 
    \text{,}
\end{equation}
where $w_c$ is the weight for class $c$. This weight is inversely proportional to the prevalence of $c$ across all EA view heatmaps in the whole dataset. For our two-class dataset, $w_c=1-\pi_c$, where $\pi_c$ is the proportion of elements labeled as $c$ across all annotations of the EA view. The resulting weights are 0.011 and 0.989 for the classes \texttt{background} and \texttt{person}, respectively. 
The soft Dice loss is given by
\begin{equation}\label{eq:SDice_loss}
    \mathcal{L}_\text{sDice} = \frac{1}{K}\sum\limits_{c \in C} \left(1 - \frac{2\sum\limits_{i=1}^M\sum\limits_{j=1}^N y_{ij}^{(c)} p_{ij}^{(c)}}{\sum\limits_{i=1}^M\sum\limits_{j=1}^N \left(y_{ij}^{(c)}\right)^2 + \sum\limits_{i=1}^M\sum\limits_{j=1}^N \left(p_{ij}^{(c)}\right)^2}\right)
    \text{.}
\end{equation}
In addition, we adapt the combination loss function: %
\begin{equation}\label{eq:loss}
    \mathcal{L}=
    \lambda_\text{wCE}\mathcal{L}_\text{wCE}+
    \lambda_\text{SDice}\mathcal{L}_\text{SDice}\text{,}
\end{equation}
with weighting factors $\lambda_\text{wCE}=1$ and $\lambda_\text{SDice}=10$, as in \cite{mvrss}.

\section{Results}
We have employed TMVA4D-CNN-5 models trained with the combination loss (\ref{eq:loss}) to provide an initial estimate of the possible segmentation performance across all dataset realizations.
Table~\ref{table:TMVA4D_CNN_performance} summarizes the architecture's performance on the test set, evaluated on the predictions in the EA view. 
From the table, we see that ``Max power'', followed by ``Mean power'' and ``Sum power'' realizations of the dataset yield the three best Intersection over Union (IoU) and Dice scores.

We have conducted ablation studies to assess the impact on model performance of removing ASPP blocks and reducing the number of input views, with results presented in Table~\ref{table:ablation}. The results demonstrate the importance of ASPP modules for high segmentation fidelity, as they capture multi-scale features while retaining spatial resolution. We further observe that incorporating elevation data mapped to the range and Doppler dimensions (as in the ER and ED views) improves performance by preserving more of the spatial-kinematic information originally represented in the point clouds.
\begin{table}[!b]
    \vspace{-0.25cm}
    \centering
    \begin{tabular}{l|c@{\hspace{20pt}}c@{\hspace{14pt}}c@{\hspace{10pt}}c}
    \toprule
    & Dice & IoU & Precision & Recall   \\
    \midrule
    
    \textbf{Max power} & 73.2 & 57.8 & 74.6 & 71.9 \\
    Mean power & 68.6 & 52.3 & 68.8 & 68.5  \\
    Sum power   & 69.0 & 52.7 & 78.7 & 61.5 \\
    Max RCS    & 63.9 & 46.9 & 66.3 & 61.7\\
    Mean RCS   & 60.5 & 43.4 & 57.7 & 63.6 \\    
    Sum RCS    & 66.4 & 49.7 & 70.7 &  62.6 \\
    
    \bottomrule
    \end{tabular}
    \caption{Segmentation performance (in \%) of TMVA4D-CNN-5 for class \texttt{person} on the test set, trained and tested on different realizations of the dataset with combination loss~(\ref{eq:loss}).}
    \label{table:TMVA4D_CNN_performance}
\end{table}
\begin{table}[!b]
    \centering
    \begin{tabular}{c|cc|cc|cl}
    \toprule
    Architecture & \multicolumn{2}{c|}{Loss: wCE} &   \multicolumn{2}{c|}{Loss: sDice} & \multicolumn{2}{c}{Loss: wCE + sDice} \\
     modification &  Dice & IoU & Dice & IoU & Dice & IoU \\
    \midrule
    w/o ASPP & 35.0 & 21.2 & 33.6 & 20.2 & 37.1  &  22.7 \\
    w/o ER \& ED & 63.3 & 46.3 & 67.7 & 51.2 & 67.0 & 50.3\\
    None & 58.8 & 41.7 & 71.7 & 55.9 & 73.2 & 57.8\\
    \bottomrule
    \end{tabular}
    \caption{Analysis of the effects of excluding ASPP modules and of omitting the ER and ED input views on the segmentation performance of TMVA4D-CNN-5 on the test set (``Max power" realization). Results are presented in percent (\%).}
    \label{table:ablation}
    \vspace{-0.15cm}
\end{table}
\begin{table*}[!b]
    \vspace{-0.06cm}
    \centering
    \begin{tabular}{ll|c@{\hspace{17pt}}c@{\hspace{10pt}}c@{\hspace{6pt}}c|c@{\hspace{17pt}}c@{\hspace{10pt}}c@{\hspace{6pt}}c|c@{\hspace{17pt}}c@{\hspace{10pt}}c@{\hspace{6pt}}c}
    \toprule
\multicolumn{1}{c}{Architecture} & Dataset & \multicolumn{4}{c}{Loss: wCE} &   \multicolumn{4}{c}{Loss: sDice} &  \multicolumn{4}{c}{Loss: wCE + sDice} \\
    
    &  & Dice & IoU & Precision & Recall & Dice & IoU & Precision & Recall & Dice & IoU & Precision & Recall \\
    \midrule
    \multicolumn{1}{c}{\textit{TMVA4D-CNN-1}} & Max  & 53.6 & 36.6  & 47.7 & 61.1  & 67.2 & 50.7 & 69.6  &  65.1 & 61.2 & 44.3 & 53.6 & 71.8\\
     &   Mean &  51.7 &  34.9 & 47.0 &  57.4 & 61.6 &  44.5 &  68.9 & 55.6 & 55.6 & 38.5  & 51.0  &   61.1 \\
    \multicolumn{1}{c}{(5.5M trainable param.)} &  Sum  & 44.4  & 28.6 & 59.5 & 35.4 & 61.5 & 44.4 & 61.6 & 61.4 & 55.3 & 38.2 & 67.3  & 47.0  \\
    \midrule
    \multicolumn{1}{c}{\textit{TMVA4D-CNN-3}} &          Max  & 64.8 & 47.9 & 61.3 & 68.7 & 70.7 & 54.7 & 81.1 & 62.7 & 72.7 & 57.2 & 73.7 & 71.8\\
     &   Mean & 55.5 & 38.4 & 59.9 & 51.6 & 69.2 & 52.9 & 81.3 & 60.2 & 65.7 & 49.0 & 65.1 & 66.4 \\
    \multicolumn{1}{c}{(5.5M trainable param.)} &  Sum  & 48.6 & 32.1 & 40.6 & 60.4 & 66.7 & 50.1 & 80.0 & 57.2 & 64.5 & 47.6 & 63.4 & 65.7 \\
    \midrule
    \multicolumn{1}{c}{\textit{TMVA4D-CNN-5}} &          Max  & 58.8 & 41.7 & 71.1 & 50.2 & 71.7 & 55.9 & 82.5 & 63.4 & 73.2 & 57.8 & 74.6 & \underline{71.9}\\
     &   Mean & 60.0 & 42.8 & 62.8 & 57.4 & 71.7 & 55.9 & 81.4 & 64.1 & 68.6 & 52.3 & 68.8 & 68.5 \\
    \multicolumn{1}{c}{(7.7M trainable param.)} &  Sum  & 58.1 & 40.9 & 58.2 & 58.0 & 66.7 & 50.0 & 83.4 & 55.6 & 69.0 & 52.7 & 78.7 & 61.5  \\
    \midrule
    \multicolumn{1}{c}{\textit{TMVA4D-CNN-7}} &          Max  & 65.4 & 48.5 & 67.6 & 63.3 & 70.2 & 54.1 & \underline{84.0} &  63.0 & 72.5 & 56.9 & 79.7 & 66.5\\
     &   Mean & 65.6 & 41.2 & 63.8 & 53.8 & 68.9 & 52.5 & 80.8 & 60.0 & 68.9 & 52.5 & 69.4 & 68.3 \\
    \multicolumn{1}{c}{(9.9M trainable param.)} &  Sum  & 58.2 & 41.1 & 58.0 & 58.5 & 70.2 & 54.1 & 84.1 & 60.3 & 67.0 & 50.4 & 65.2 & 69.0 \\
    \midrule
    \multicolumn{1}{c}{\textit{TMVA4D-ConvLSTM}}        & Max  & 65.5 & 48.7 & 60.6 & 71.2 & 73.6 & 58.3 & 80.6 & 67.8 & 72.2 & 56.5 & 79.5 & 66.1  \\
    & Mean & 62.0 & 45.0 & 72.0 & 54.5 & 73.0 & 57.5 & 81.5 & 66.1 & 71.3 & 55.4 & 77.0 & 66.4 \\
   \multicolumn{1}{c}{(133M trainable param.)} & Sum  & 65.8 & 49.0 & 62.6 & 69.3 & 65.9 & 49.1 & 83.3 & 54.5 & 68.8 & 52.5 & 79.9 & 60.5 \\
    \midrule
    \multicolumn{1}{c}{\textit{TMVA4D-IntConvLSTM}}&    Max & 63.3 & 46.3 & 59.4 & 67.6 & 74.8 & 59.7 & 83.9 & 67.5 & 71.2 & 55.4 & 75.5 & 67.5 \\
    \multicolumn{1}{c}{``Light"} &   Mean  & 58.5 & 41.3 & 57.8 & 59.1 & 70.0 & 53.9 & 81.0 & 61.7 & 72.5 & 56.9 & 71.8 & \textbf{73.3} \\
    \multicolumn{1}{c}{(65M trainable param.)} &  Sum & 45.8 & 29.7 & 35.1 & 65.8  & 67.4 & 51.0 & 70.5 & 64.7 & 69.4 & 53.2 & 66.5 & 72.6 \\

    \midrule
    \multicolumn{1}{c}{\textit{TMVA4D-IntConvLSTM}}&    Max & 62.6 & 45.6  & 58.0 & 68.0 & \textbf{75.9} & \textbf{61.2} &  \textbf{84.3} & 69.1  & 72.5 &  56.9 & 74.7  & 70.5 \\
    \multicolumn{1}{c}{``Medium"} &   Mean  & 55.7 &  38.6 & 58.0 &  53.6 &  73.7 &  58.4 & 83.5  &  66.1 & 72.4  & 56.6  &  77.7 &  67.5  \\
    \multicolumn{1}{c}{(97.7M trainable param.)} &  Sum   &  52.9 &  36.0 & 62.7  & 45.8  &  67.7 &  51.2 & 80.7  & 58.4 & 70.8  & 54.8  &  78.2 & 64.7 \\
    
    \midrule
    \multicolumn{1}{c}{\textit{TMVA4D-IntConvLSTM}}&    Max & 56.2 & 39.1 & 53.1 & 59.8 & \underline{75.6} & \underline{60.7} & 83.0 & 69.3 & 73.0 & 57.5 & 80.1 & 67.1 \\
    \multicolumn{1}{c}{``Heavy"} &   Mean  & 54.8 & 37.8 & 48.2 & 63.6 & 71.0 & 55.0 & 80.8 & 63.2 & 65.0 & 48.1  & 73.7  &  58.1 \\
    \multicolumn{1}{c}{(132M trainable param.)} &      Sum & 50.9 & 34.1 & 43.0 & 62.3 & 66.8 & 50.2 & 82.1 & 56.4 & 70.4 & 54.3 & 71.4 & 69.5 \\
    
    \bottomrule
    \end{tabular}
    \caption{Segmentation performance (in \%) of TMVA4D models with TMVA4D-CNN, TMVA4D-ConvLSTM, and TMVA4D-IntConvLSTM encoders for class \texttt{person} on the test set for ``Max power'', ``Mean power'', and ``Sum power'' realizations of the dataset with wCE (\ref{eq:wCE_loss}), sDice (\ref{eq:SDice_loss}), and wCE + sDice (\ref{eq:loss}) losses. Best results for each metric are shown in bold and second best results are underlined. %
    }
    \label{table:TMVA4D_ALL_performance}
\end{table*}

As our method is the first that uses five input views, including elevation data, to perform semantic segmentation of 2D radar heatmaps, direct comparisons with previous work are infeasible. The closest prior method, TMVA-Net~\cite{mvrss}, is analogous to TMVA4D-CNN-5 without the ER and ED views, with our results (see Table~\ref{table:ablation}) showing the performance benefit of leveraging elevation data.

Next, on the "power" dataset realizations we evaluate the performance of the TMVA4D-CNN-$\{ i \}$ ($i \in \{1,3,5,7\}$) models alongside the TMVA4D-ConvLSTM and TMVA4D-IntConvLSTM models, which incorporate ConvLSTM layers. For model training, we use the combination loss (\ref{eq:loss}) as well as its individual components: weighted cross-entropy loss (\ref{eq:wCE_loss}) and soft Dice loss (\ref{eq:SDice_loss}). We summarize the results achieved using these models and loss functions in Table~\ref{table:TMVA4D_ALL_performance}.

As shown in Table~\ref{table:TMVA4D_ALL_performance}, the smallest non-recurrent model to take multiple frames of input (TMVA4D-CNN-3) achieves respectable segmentation performance for class \texttt{person}, with a Dice score of 72.7\% and IoU of 57.2\% on the ``Max power'' dataset and combination loss (\ref{eq:loss}). This constitutes a significant improvement over using a single input frame (TMVA4D-CNN-1), which does not incorporate temporal information and achieves Dice and IoU scores of 67.2\% and 50.7\%, respectively.
Increasing the number of input frames to five (TMVA4D-CNN-5) modestly improves the segmentation performance to 73.2\% Dice and 57.8\% IoU scores, whereas increasing it to seven (TMVA4D-CNN-7) does not bring further improvements. We note that the five-frame input adds one 3D CNN layer, while the seven-frame input adds two 3D CNN layers to the encoders compared to the three-frame input model. Additional refinement of spatiotemporal representation learning is achieved using ConvLSTM encoders. These enhance the modeling of temporal dependencies through ConvLSTM layers applied to the patterns extracted by 3D CNNs in the TMVA4D-ConvLSTM encoders. The result shows a minor improvement of $\sim$\,0.5\% compared to TMVA4D-CNN-5 (both using five input frames). TMVA4D-IntConvLSTM, which interleaves ConvLSTM and 3D CNN layers to enhance space-time feature learning, achieves the highest segmentation performance, with Dice 74.8\%, 75.9\%, and 75.6\% (IoU: 59.7\%, 61.2, and 60.7\%) for the ``Light”, ``Medium”, and ``Heavy” versions, respectively.

\definecolor{col1}{HTML}{3282be}
\definecolor{col2}{HTML}{f7a609}
\definecolor{col3}{HTML}{be281b}
\definecolor{col4}{HTML}{006f41}
\definecolor{col5}{HTML}{133455}
\definecolor{col6}{HTML}{800080}
\begin{figure*}[!b]
    \vspace{-0.08cm}
    \centering
    \begin{subfigure}{0.245\textwidth}%
      \begin{tikzpicture}[]
        \node[anchor=south west,inner sep=0] (image) at (0,0)
        {\includegraphics[clip,trim=0mm 0mm 0mm 45mm,width=\linewidth,height=0.60\linewidth]{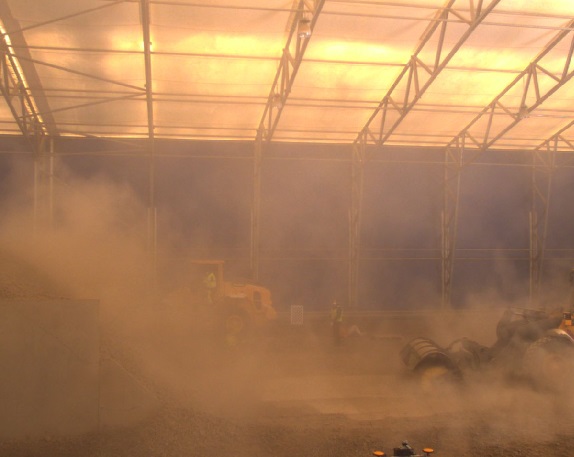}};
        \begin{scope}[x={(image.south east)},y={(image.north west)}]
          \draw[col1,] (0.3425,0.46) rectangle++ (0.035,0.21);
          \draw[col2,] (0.385,0.37) rectangle++ (0.0725,0.09);
          \draw[col3,] (0.5625,0.37) rectangle++ (0.0335,0.19);
          \draw[col4,] (0.57,0.4) rectangle++ (0.065,0.09);
        \end{scope}
      \end{tikzpicture}\\
      \begin{tikzpicture}[]
        \node[anchor=south west,inner sep=0] (image) at (0,0)
        {\includegraphics[clip,trim=0mm 0mm 0mm 70mm,width=\linewidth,height=0.60\linewidth]{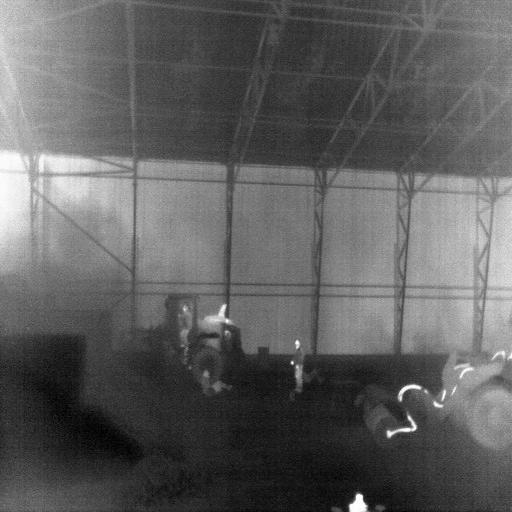}};
        \begin{scope}[x={(image.south east)},y={(image.north west)}]
          \draw[col1,] (0.3425,0.46) rectangle++ (0.035,0.21);
          \draw[col2,] (0.385,0.37) rectangle++ (0.0725,0.09);
          \draw[col3,] (0.5625,0.37) rectangle++ (0.0335,0.19);
          \draw[col4,] (0.57,0.4) rectangle++ (0.065,0.09);
        \end{scope}
      \end{tikzpicture}\\
      \begin{tikzpicture}[]
        \node[anchor=south west,inner sep=0] (image) at (0,0)
        {\includegraphics[clip,trim=0mm 0mm 0mm 13mm,width=\linewidth,height=0.60\linewidth]{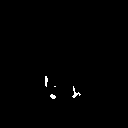}};
        \begin{scope}[x={(image.south east)},y={(image.north west)}]
          \draw[col1,] (0.3425,0.46) rectangle++ (0.035,0.21);
          \draw[col2,] (0.385,0.37) rectangle++ (0.0725,0.09);
          \draw[col3,] (0.5625,0.37) rectangle++ (0.0335,0.19);
          \draw[col4,] (0.57,0.4) rectangle++ (0.065,0.09);
        \end{scope}
      \end{tikzpicture}\\
        \vspace{-15pt}
        \caption{Tent -- Stationary -- Dust}
    \end{subfigure}
    \hfill
    \begin{subfigure}{0.245\textwidth}
      \begin{tikzpicture}[]
        \node[anchor=south west,inner sep=0] (image) at (0,0)
        {\includegraphics[clip,trim=0mm 3.5mm 1.5mm 49mm,width=\linewidth,height=0.60\linewidth]{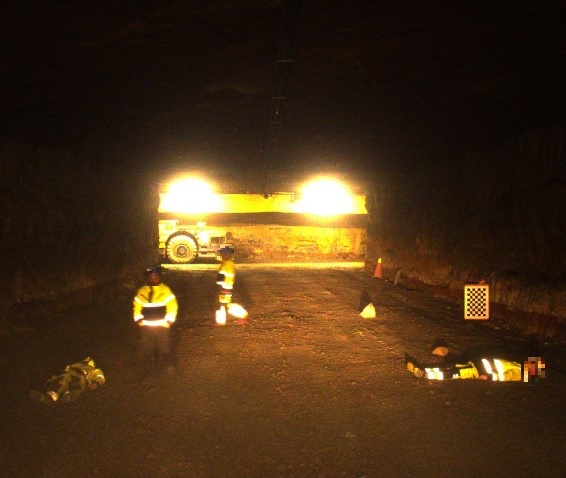}};
        \begin{scope}[x={(image.south east)},y={(image.north west)}]
          \draw[col1,] (0.05,0.22) rectangle++ (0.15,0.18);
          \draw[col2,] (0.24,0.29) rectangle++ (0.08,0.45);
          \draw[col3,] (0.38,0.49) rectangle++ (0.05,0.31);
          \draw[col4,] (0.72,0.29) rectangle++ (0.245,0.09);
        \end{scope}
      \end{tikzpicture}\\
      \begin{tikzpicture}[]
        \node[anchor=south west,inner sep=0] (image) at (0,0)
        {\includegraphics[clip,trim=0mm 0mm 0mm 70mm,width=\linewidth,height=0.60\linewidth]{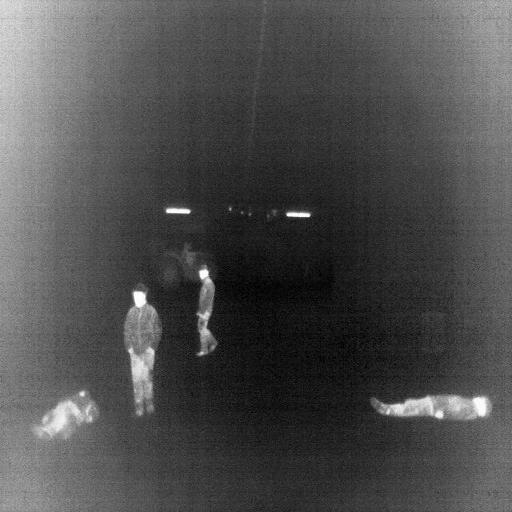}};
        \begin{scope}[x={(image.south east)},y={(image.north west)}]
          \draw[col1,] (0.05,0.22) rectangle++ (0.15,0.18);
          \draw[col2,] (0.24,0.29) rectangle++ (0.08,0.45);
          \draw[col3,] (0.38,0.49) rectangle++ (0.05,0.31);
          \draw[col4,] (0.72,0.29) rectangle++ (0.245,0.09);
        \end{scope}
      \end{tikzpicture}\\
      \begin{tikzpicture}[]
        \node[anchor=south west,inner sep=0] (image) at (0,0)
        {\includegraphics[clip,trim=0mm 0mm 0mm 13mm,width=\linewidth,height=0.60\linewidth]{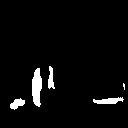}};
        \begin{scope}[x={(image.south east)},y={(image.north west)}]
          \draw[col1,] (0.05,0.22) rectangle++ (0.15,0.18);
          \draw[col2,] (0.24,0.29) rectangle++ (0.08,0.45);
          \draw[col3,] (0.38,0.49) rectangle++ (0.05,0.31);
          \draw[col4,] (0.72,0.29) rectangle++ (0.245,0.09);
        \end{scope}
      \end{tikzpicture}\\
        \vspace{-15pt}
        \caption{Mine -- Stationary -- Mist}
    \end{subfigure}
    \hfill
    \begin{subfigure}{0.245\textwidth}
      \begin{tikzpicture}[]
        \node[anchor=south west,inner sep=0] (image) at (0,0)
        {\includegraphics[clip,trim=1.8mm 2mm 1.9mm 54mm,width=\linewidth,height=0.60\linewidth]{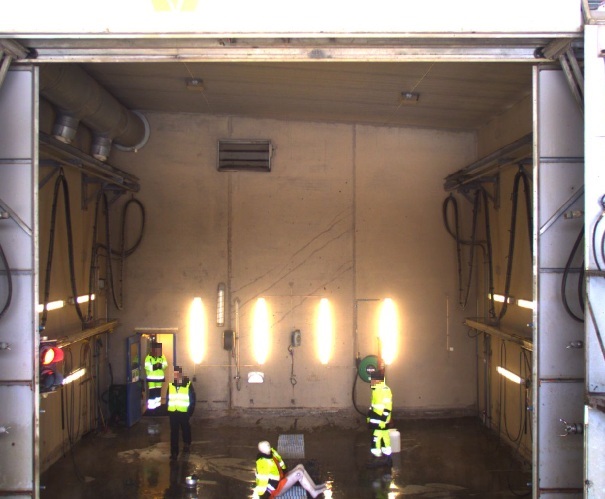}};
        \begin{scope}[x={(image.south east)},y={(image.north west)}]
          \draw[col1,] (0.232,0.27) rectangle++ (0.045,0.26);
          \draw[col2,] (0.267,0.11) rectangle++ (0.057,0.337);
          \draw[col3,] (0.42,0.01) rectangle++ (0.125,0.17);
          \draw[col4,] (0.6,0.075) rectangle++ (0.056,0.35);
        \end{scope}
      \end{tikzpicture}\\
      \begin{tikzpicture}[]
        \node[anchor=south west,inner sep=0] (image) at (0,0)
        {\includegraphics[clip,trim=0mm 0mm 0mm 70mm,width=\linewidth,height=0.60\linewidth]{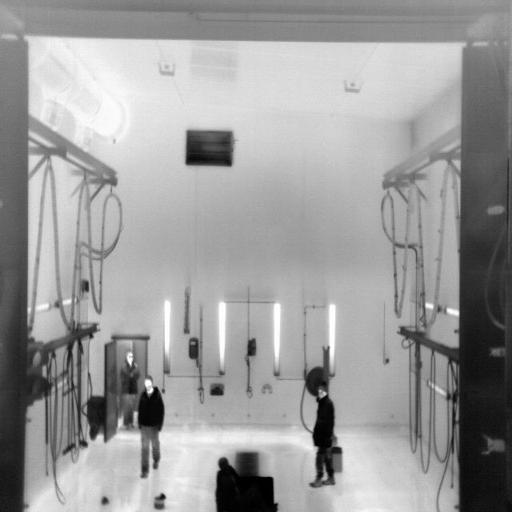}};
        \begin{scope}[x={(image.south east)},y={(image.north west)}]
          \draw[col1,] (0.232,0.27) rectangle++ (0.045,0.26);
          \draw[col2,] (0.267,0.11) rectangle++ (0.057,0.337);
          \draw[col3,] (0.42,0.01) rectangle++ (0.125,0.17);
          \draw[col4,] (0.6,0.075) rectangle++ (0.056,0.35);
        \end{scope}
      \end{tikzpicture}\\
      \begin{tikzpicture}[]
        \node[anchor=south west,inner sep=0] (image) at (0,0)
        {\includegraphics[clip,trim=0mm 0mm 0mm 13mm,width=\linewidth,height=0.60\linewidth]{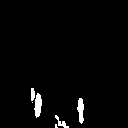}};
        \begin{scope}[x={(image.south east)},y={(image.north west)}]
          \draw[col1,] (0.232,0.27) rectangle++ (0.045,0.26);
          \draw[col2,] (0.267,0.11) rectangle++ (0.057,0.337);
          \draw[col3,] (0.42,0.01) rectangle++ (0.125,0.17);
          \draw[col4,] (0.6,0.075) rectangle++ (0.056,0.35);
        \end{scope}
      \end{tikzpicture}\\
        \vspace{-15pt}
        \caption{Car wash -- Stationary -- Clear}
        \label{subfig:mine_stationary_prediction}
    \end{subfigure}
    \hfill
    \begin{subfigure}{0.245\textwidth}
      \begin{tikzpicture}[]
        \node[anchor=south west,inner sep=0] (image) at (0,0)
        {\includegraphics[clip,trim=15mm 10mm 18mm 43mm,width=\linewidth,height=0.60\linewidth]{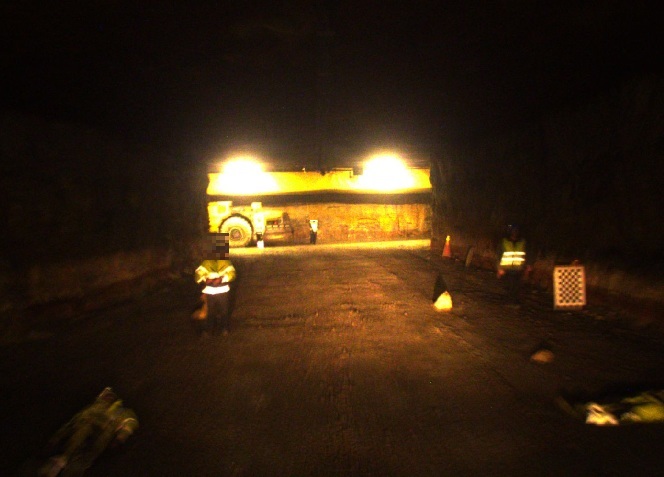}};
        \begin{scope}[x={(image.south east)},y={(image.north west)}]
          \draw[col1,] (0.007,0.01)rectangle++ (0.135,0.18);
          \draw[col2,] (0.255,0.35) rectangle++ (0.075,0.39);
          \draw[col3,] (0.803,0.46) rectangle++ (0.075,0.34);
          \draw[col4,] (0.37,0.692) rectangle++ (0.025,0.125);
          \draw[col5,] (0.467,0.705) rectangle++ (0.022,0.11);
          \draw[col6,] (0.93,0.02) rectangle++ (0.062,0.135);
        \end{scope}
      \end{tikzpicture}\\
      \begin{tikzpicture}[]
        \node[anchor=south west,inner sep=0] (image) at (0,0)
        {\includegraphics[clip,trim=0mm 0mm 0mm 70mm,width=\linewidth,height=0.60\linewidth]{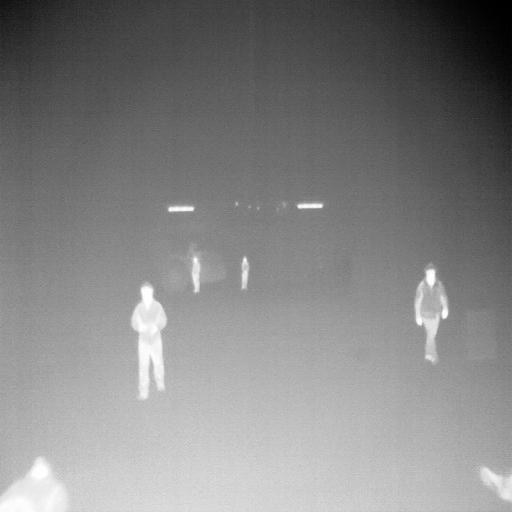}};
        \begin{scope}[x={(image.south east)},y={(image.north west)}]
          \draw[col1,] (0.007,0.01)rectangle++ (0.135,0.18);
          \draw[col2,] (0.255,0.35) rectangle++ (0.075,0.39);
          \draw[col3,] (0.803,0.46) rectangle++ (0.075,0.34);
          \draw[col4,] (0.37,0.692) rectangle++ (0.025,0.125);
          \draw[col5,] (0.467,0.705) rectangle++ (0.022,0.11);
          \draw[col6,] (0.93,0.02) rectangle++ (0.062,0.135);
        \end{scope}
      \end{tikzpicture}\\
      \begin{tikzpicture}[]
        \node[anchor=south west,inner sep=0] (image) at (0,0)
        {\includegraphics[clip,trim=0mm 0mm 0mm 13mm,width=\linewidth,height=0.60\linewidth]{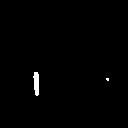}};
        \begin{scope}[x={(image.south east)},y={(image.north west)}]
          \draw[col1,] (0.007,0.01)rectangle++ (0.135,0.18);
          \draw[col2,] (0.255,0.35) rectangle++ (0.075,0.39);
          \draw[col3,] (0.803,0.46) rectangle++ (0.075,0.34);
          \draw[col4,] (0.37,0.692) rectangle++ (0.025,0.125);
          \draw[col5,] (0.467,0.705) rectangle++ (0.022,0.11);
          \draw[col6,] (0.93,0.02) rectangle++ (0.062,0.135);
        \end{scope}
      \end{tikzpicture}\\
        \vspace{-15pt}
        \caption{Mine -- Ego-motion -- Clear}
        \label{subfig:mine_moving_prediction}
    \end{subfigure}
    \caption{Images and corresponding TMVA4D-CNN-5 predictions for frames in the test set ("Max power" realization). Each subfigure shows for a particular frame its color camera image (top), thermal camera image (middle), and radar-only semantic segmentation (bottom). Predictions show class \texttt{background} in black and class \texttt{person} in white. Subcaptions lists the environment, presence of ego-motion, and visibility condition in the frame. Bounding boxes (drawn manually, for visualization only) indicate the same person in each column.
    }
    \label{fig:qualitative_cropped}
\end{figure*}

As can be seen in Table~\ref{table:TMVA4D_ALL_performance}, the best segmentation performance for all architectures was achieved on the ``Max power'' realization of the dataset. The weighted cross-entropy loss (\ref{eq:wCE_loss}) results in the overall worst performance for all architectures since it computes a pixel-wise loss, treating each pixel independently without optimizing for overlap quality.
For architectures that utilize ConvLSTM layers in the encoder, the best performance is achieved for models trained using the sDice loss function (\ref{eq:SDice_loss}), which encourages holistic and coherent object shapes. The combination loss (\ref{eq:loss}) yields the best results for the TMVA4D-CNN architectures, overcoming the difficulties when training with the sDice loss (\ref{eq:SDice_loss}) (as previously shown by \cite{mvrss} for TMVA-Net).

Qualitative results achieved by TMVA4D-CNN-5 are shown in Fig.~\ref{fig:qualitative_cropped}. We find that model performance between settings varies in several ways: the models more often fail to detect individuals lying on the ground, while people in motion are easily detected; predictions are also less accurate for frames recorded in ego-motion and for people located far from the sensors. We identify multiple factors that likely contribute to these disparities.
First, for people lying down, the low resolution of the radar (compared to lidar) is a limiting factor in representing such people as separate from the ground. The projection from point clouds to heatmaps exacerbates this issue, as spatial information is lost in the process. Furthermore, the intensities of radar return signals corresponding to such people are often similar to those from the ground. Second, motionless people cannot be distinguished from their environment by means of Doppler measurements. Finally, there is an imbalance in the dataset, with people recorded mostly from a stationary vehicle.

In Table~\ref{table:environment-performance} we present the segmentation performance achieved by the ``Medium" TMVA4D-IntConvLSTM model in contrasting environments (``Mine" and ``Surface") and visibility conditions (with and without dust). The results indicate that airborne particles have little impact on performance. Differences between settings are likely due to variations in ego-motion, person count, and sensor-to-person distances.
\setcounter{table}{3}
\begin{table}[!b]
    \vspace{-0.3cm}
    \centering
    \begin{tabular}{cc|cc??cc|cc}
    \toprule
    \multicolumn{2}{c|}{Mine} & \multicolumn{2}{c??}{Surface} &   \multicolumn{2}{c|}{Clear} & \multicolumn{2}{c}{Dust} \\
        Dice & IoU  &  Dice & IoU & Dice & IoU & Dice & IoU \\
    \midrule
    79.6   &  66.1 & 68.3 & 51.9 & 73.2 & 57.8 & 71.3  &  55.4 \\      
    \bottomrule
    \end{tabular}
    \caption{Segmentation performance (in \%) of the ``Medium" TMVA4D-IntConvLSTM architecture trained using sDice loss on the different subsets of the test set (``Max power" realization). The "Clear" and "Dust" conditions correspond to the ``Tent'' environment, which is a subset of the ``Surface'' (non-mine) environment.}
    \label{table:environment-performance}
    \vspace{-0.1cm}
\end{table}

In addition to evaluating TMVA4D models on our test set, we have also deployed TMVA4D-CNN-5 for live demonstrations. These experiments demonstrate the viability of our approach for real-time deployment, achieving an output frequency above 8 Hz for point clouds with fewer than 5000 points using an NVIDIA GeForce RTX 2080 Ti GPU.%

\section{Conclusions}

In this work, we introduce a dataset for human detection designed for field robotics applications with a focus on 4D radar. The dataset features humans in various positions and modes of motion in a range of field environments (mining, construction) and visibility conditions (dust, water, smoke). %

We also present TMVA4D, a family of deep learning artificial neural network architectures for high-performance semantic segmentation of people in 4D radar data. TMVA4D processes multiple frames of radar heatmaps across five 2D projections from 4D radar point clouds, predicting semantic segmentation masks in the elevation-azimuth view.

We have evaluated TMVA4D on our 4D radar dataset, demonstrating its high performance for human detection in field environments. The ``Medium" TMVA4D-IntConvLSTM version provides the best quantitative segmentation performance. However, the TMVA4D-CNN models establish non-recurrent convolutional architectures as a compelling option for live robot deployment, achieving comparable performance at lower computational cost due to their smaller size.

\printbibliography

\end{document}